\documentclass[sigconf]{acmart}

\usepackage{times}
\usepackage{epsfig}
\usepackage{dsfont}
\usepackage{amsmath}
\usepackage{amssymb}
\usepackage{graphicx}
\usepackage{enumitem}
\usepackage{dirtytalk}
\usepackage{mathtools}
\usepackage{tikz}
\usepackage[caption=false]{subfig}
\usepackage[export]{adjustbox}
\usepackage{array}
\usepackage{rotating}
\usepackage{multirow}
\usepackage{colortbl}
\usepackage{soul}

\setcopyright{acmlicensed}

\newcommand{\FF}{\mathcal{F}}

\newcommand{\Bat}{\mathbb{B}}

\newcommand{\iqry}{{q}}
\newcommand{\Pos}{\mathbb{P}}
\newcommand{\ipos}{{p}}

\newcommand{\ineg}{{n}}

\newcommand{\Loss}{\mathcal{L}}
\newcommand{\loss}{\ell}
\newcommand{\isem}{{s}}
\newcommand{\iret}{{r}}

\acmConference[SIGIR'18]{The 41st International ACM SIGIR Conference on Research and Development in Information Retrieval, Ann Arbor, MI, USA}
\acmYear{2018}
\copyrightyear{2018}
\acmArticle{4}
\acmPrice{15.00}

\newcommand{\micael}[1]{{#1}}

\renewcommand{\textrightarrow}{$\rightarrow$}

\begin{document}
\begin{sloppypar}
    \title{
        Cross-Modal~Retrieval~in~the~Cooking~Context:
        Learning~Semantic~Text-Image~Embeddings
    }
    
    \author{Micael Carvalho}
    \authornote{Equal contribution}
    \affiliation{Sorbonne Universit\'e, CNRS, LIP6, F-75005 Paris, France}
    \email{micael.carvalho@lip6.fr}
    
    \author{R\'emi Cad\`ene}
    \authornotemark[1]
    \affiliation{Sorbonne Universit\'e, CNRS, LIP6, F-75005 Paris, France}
    \email{remi.cadene@lip6.fr}
    
    \author{David Picard}
    \affiliation{Sorbonne Universit\'e, CNRS, LIP6, F-75005 Paris, France}
    \email{david.picard@lip6.fr}
    
    \author{Laure Soulier}
    \affiliation{Sorbonne Universit\'e, CNRS, LIP6, F-75005 Paris, France}
    \email{laure.soulier@lip6.fr}
    
    \author{Nicolas Thome}
    \affiliation{CEDRIC - Conservatoire National des Arts et M\'etiers, F-75003 Paris, France}
    \email{nicolas.thome@cnam.fr}
    
    \author{Matthieu Cord}
    \affiliation{Sorbonne Universit\'e, CNRS, LIP6, F-75005 Paris, France}
    \email{matthieu.cord@lip6.fr}

    \begin{abstract}
        
        Designing powerful tools  that support cooking activities has rapidly gained popularity due to the massive amounts of available data, as well as recent advances in machine learning that are capable of analyzing them. 
        In this paper, we propose a cross-modal retrieval model aligning visual and textual data (like pictures of dishes and their recipes) in a shared representation space.
        We describe an effective learning scheme, capable of tackling large-scale problems, and validate it on the Recipe1M dataset containing nearly 1 million picture-recipe pairs.
        We show the effectiveness of our approach regarding previous state-of-the-art models and present qualitative results over computational cooking use cases.
        
    \end{abstract}
    
    \ccsdesc[500]{Information systems~Multimedia and multimodal retrieval}
    \ccsdesc[500]{Computer systems organization~Neural networks}
    
    \keywords{Deep Learning, Cross-modal Retrieval, Semantic Embeddings}
    
    \maketitle
    
    \section{Introduction}
    Designing powerful tools that support cooking activities has become an attractive research field in recent years due to the growing interest of users to eat home-made food and share recipes on social platforms \cite{SanjoK17}.
These massive amounts of data shared on devoted sites, such as \textit{All Recipes}\footnote{\url{http://www.allrecipes.com/}}, allow  gathering food-related data including text recipes, images, videos, and/or user preferences. Consequently, novel applications are rising, such as ingredient classification~\cite{chen2016deep}, recipe recognition~\cite{wang2015} or recipe recommendation~\cite{SanjoK17}. 
However, solving these tasks is challenging since it requires taking into consideration 1) the heterogeneity of data in terms of format (text, image, video, ...) or structure (\textit{e.g.}, list of items for ingredients, short verbal sentence for instructions, or verbose text for users' reviews); and 2) the cultural factor behind each recipe since the vocabulary, the quantity measurement, and the flavor perception is culturally  intrinsic; preventing the homogeneous semantics of recipes.

One recent approach emerging from the deep learning community aims at learning the semantics of objects in a latent space using the \textit{distributional hypothesis} \cite{Harris54} that constrains object with similar meanings to be represented similarly.  First used for learning image representations (also called \textit{embeddings}), this approach has been derived to text-based applications, and recently some researchers investigate the potential of representing multi-modal evidence sources (\textit{e.g.}, texts and images) in a shared latent space \cite{karpathy2015deep,lazaridou2015}. This research direction is particularly interesting for grounding language with common sense information extracted from images, or vice-versa. In the context of computer-aided cooking, we believe that this multi-modal representation learning approach would contribute to solving the heterogeneity challenge, since they would promote a better understanding of each domain-specific word/image/video. In practice, a typical approach consists in aligning text and image representations in a shared latent space in which they can be compared ~\cite{bossard2014,KawanoY2014,kiros2015unifying,karpathy2015deep,salvador2017,jingjing2017}.   One direct application in the cooking context is to perform cross-modal retrieval where the goal is to retrieve images similar to a text recipe query  or conversely text recipes similar to a  image query. 

\begin{figure}[tb]
    \centering
    \subfloat[im2recipe~\cite{salvador2017}]{
        \includegraphics[width=0.49\linewidth]{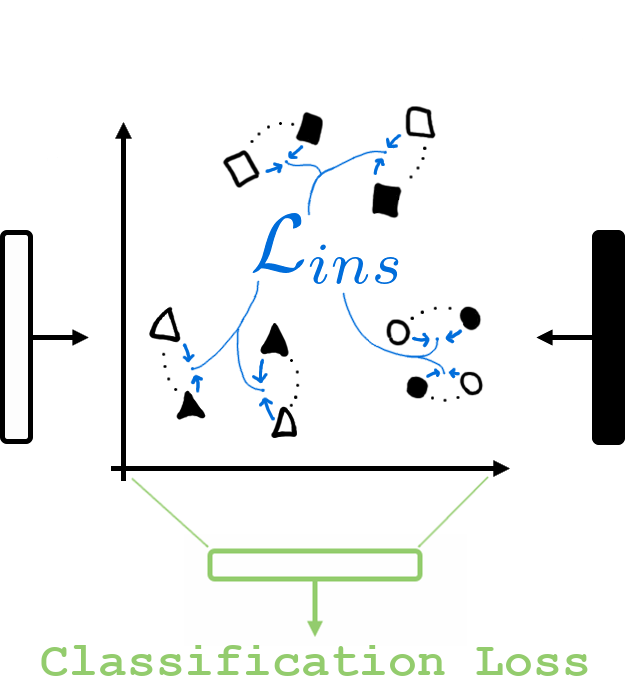}
        \label{fig:sem_loss_a}
    }
    \subfloat[AdaMine (ours)]{
        \includegraphics[width=0.49\linewidth]{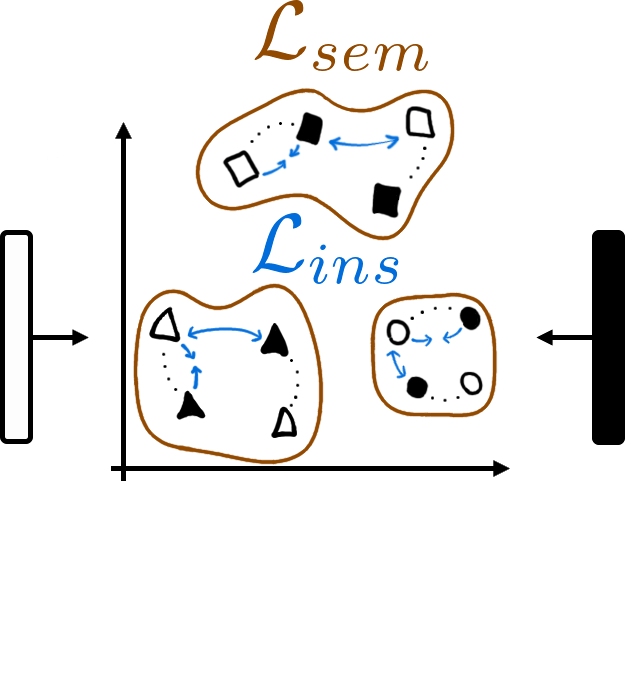}
        \label{fig:sem_loss_b}
    }
    \vspace{-0.1cm}
    \caption{Comparison between (a) the classification augmented latent space learning of \cite{salvador2017}  and  (b) our joint retrieval and semantic latent space learning, which combines instance-based ($\mathcal{L}_{ins}$) and semantic-based ($\mathcal{L}_{sem}$) losses.}
    \label{fig:sem_loss}
    \vspace{-0.6cm}
\end{figure}

However, Salvador et al.  \cite{salvador2017} highlight that this solution based on aligning matching pairs can lead to poor retrieval performances in a large scale framework. Training the latent space by only matching pairs of the exact same dish is not particularly effective at mapping similar dishes close together, which induces a lack of generalization to newer items (recipes or images). To alleviate this problem, \cite{salvador2017}  proposes to use additional data (namely, categories of meals) to train a classifier with the aim of regularizing the latent space embeddings (Figure~\ref{fig:sem_loss_a}). \micael{Their approach involves adding an extra layer to a deep neural network, specialized in the classification task.} However, we believe that such classification scheme is under-effective for two main reasons. First, the classifier adds many parameters that are discarded at the end of the training phase, since classification is not a goal of the system. Second, we hypothesize that given its huge number of parameters, the classifier can be trained with high accuracy without changing much of the structure of the underlying latent space, which completely defeats the original purpose of adding classification information.

To solve these issues, we propose a unified learning framework in which we simultaneously leverage retrieval and class-guided features in a shared latent space (Figure~\ref{fig:sem_loss_b}). 
Our contributions are three-fold:\\
\indent $\bullet$ We formulate a joint objective function with cross-modal retrieval and classification loss to structure the latent space. Our intuition is that directly injecting the class-based evidence sources in the representation learning process is more effective at enforcing a high-level structure to the latent space as shown in Figure \autoref{fig:sem_loss_b}.\\
\indent $\bullet$  We propose a double-triplet scheme to express jointly 1) the retrieval loss (e.g., corresponding picture and recipe of a pizza should be closer in the latent space than any other picture - see blue arrows in Figure \autoref{fig:sem_loss_b}) and 2) the class-based one \micael{(\textit{e.g.}, any 2 pizzas should be closer in the latent space than a pizza and another item from any other class, like salad). This double-loss is capable of taking into consideration both the fine-grained and the high-level semantic information underlying recipe items. Contrary to the proposal of \cite{salvador2017}, our class-based loss acts directly on the feature space, instead of adding a classification layer to the model}.\\
\indent $\bullet$ We introduce a new scheme to tune the gradient update in the stochastic gradient descent and back-propagation algorithms used for training our model. More specifically, we improve over the usual gradient averaging update by performing an adaptive mining of the most significant triplets, which leads to better embeddings.

We instantiate these contributions through a dual deep neural network.
We thoroughly evaluate our model on Recipe1M~\cite{salvador2017}, the only English large-scale cooking dataset available, and show its superior performances compared to the state of the art models.

The remaining of this paper is organized as follows. In Section 2, we present previous work related to computer-aided cooking and cross-modal retrieval. Then, we present in Section 3 our  joint retrieval and classification model as well as our adaptive triplet mining scheme to train the model. We introduce the experimental protocol in Section 4. In Section 5 we experimentally validate our hypotheses and present quantitative and qualitative results highlighting our model effectiveness.
Finally, we conclude and discuss perspectives.

    \section{Related work}
    \subsection{Computational cooking}
Cooking is one of the most fundamental human activities connected to various aspects of human life such as food, health, dietary, culinary art, and so on.
This is more particularly perceived on social platforms in which people share recipes or their opinions about meals, also known as the \textit{eat and tweet} or \textit{food porn} phenomenon  \cite{Amato2017}. Users' needs give rise to  smart cooking-oriented tasks 
that contribute towards the definition of computational cooking as a research field by itself \cite{cea2017}. 
Indeed, the research community is very active in investigating issues regarding food-related tasks, such as ingredient identification~\cite{chen2016}, recipe recommendation \cite{Elsweiler2017}, or recipe popularity prediction~\cite{SanjoK17}.  A first line of work consists in leveraging the semantics behind recipe texts and images using deep learning approaches \cite{salvador2017,SanjoK17}. The objective of such propositions is to generally align different modalities in a shared latent space to perform cross-modal retrieval or recommendation.  For instance, Salvador et al. \cite{salvador2017}  introduce a dual neural network that aligns textual and visual representations under both the  distributional hypothesis and classification constraints  \cite{salvador2017}. A second line of work  \cite{Elsweiler2017,Trattner2017,Kusmierczyk2016} aims at exploiting additional information (\textit{e.g.}, calories, biological or economical factors) to bridge the gap between computational cooking and  healthy issues. For instance, Kusmierczyk et al. \cite{Kusmierczyk2016} extend the Latent Dirichlet Algorithm (LDA) for combining recipe descriptions and nutritional related meta-data to mine  latent recipe topics that could be exploited to predict nutritional values of meals.

These researches are boosted by the release of food-related datasets \cite{bossard2014,chenpfid2009icip,Farinella2015,Kawano2014}. As a first example, \cite{chenpfid2009icip} proposes the Pittsburgh fast-food image dataset, containing 4,556 pictures of fast-food plates.
In order to solve more complex tasks, other initiatives provide richer sets of images \cite{bossard2014,chen2016,Harashima2017,salvador2017,wang2015}. For example, \cite{bossard2014} proposes the Food-101 dataset, containing around 101,000 images of 101 different categories. 
Additional meta-data information is also provided in the dataset of \cite{Beijbom2015} which involves  GPS data or  nutritional values.
More recently, two very large-scale food-related datasets, respectively in English and Japanese, are released  by \cite{salvador2017} and \cite{Harashima2017}. The  \textit{Cookpad} dataset  \cite{Harashima2017} gathers more than 1 million of recipes   described using structured information, such as recipe description, ingredients, and process steps as well as images.  Salvador et al. \cite{salvador2017} have  collected a very large dataset with nearly 1 million recipes, with about associated 800,000 images. They also add extra information corresponding to recipe classes and show how this new semantic information may be helpful to improve deep cross-modal retrieval systems.
To the best of our knowledge, this dataset \cite{salvador2017} is the only large-scale English one including  a huge pre-processing step for cleaning and formatting information. The strength of this dataset is that it is composed of structured ingredients and instructions, images, and a large number of classes as well. Accordingly, we focus all our experimental evaluation on this dataset.

\vspace{-0.1cm}
\subsection{Cross-modal Retrieval}
Cross-modal retrieval aims at retrieving relevant items that are of different nature with respect to the query format; for example when querying an image dataset with keywords (image vs. text) \cite{salvador2017}.
The main challenge is to  measure the similarity between different modalities of data.
In the Information Retrieval (IR) community, early work have circumvented this issue by annotating  images  to perceive their underlying semantics \cite{Jeon2003,Sun2011}. However, these approaches generally require a supervision from users to annotate at least a small sample of images.
An unsupervised solution has emerged from the deep learning community which consists in mapping images and texts into a shared latent space $\FF$ in which they can be compared~\cite{wu17sigir}.
In order to align the text and image manifolds in $\FF$, the most popular strategies are based either on 1) global alignment methods aiming at mapping each modal manifold in $\FF$ such that semantically similar regions share the same directions in $\FF$; 2) local metric learning approaches aiming at mapping each modal manifold such that semantically similar items have a short distances in $\FF$.

In the first category of works dealing with global alignment methods, a well-known state-of-the-art model is provided by the Canonical Correlation Analysis (CCA)~\cite{hotelling1936relations} which aims at maximizing the correlation in $\FF$ between relevant pairs from data of different modalities. CCA and its variations like Kernel-CCA~\cite{lai2000kernel,bach02jmlr} and Deep-CCA~\cite{andrew13jmlr} have been successfully applied to align text and images~\cite{yan15cvpr}.
However, global alignment strategies such as CCA do not take into account dissimilar pairs, and thus tend to produce false positives in cross-modal retrieval tasks.

In the second category of work, local metric learning approaches consider cross-modal retrieval as a ranking problem where items are ranked according to their distance to the query in the latent space.
A perfect retrieval corresponds to a set of inequalities in which the distances between the query and relevant items are smaller than the distances between the query and irrelevant items \cite{weinbergerlmnn2009, xingnips2002, law13cvpr}.
Each modal projection is then learned so as to minimize a loss function that measures the cost of violating each of these inequalities \cite{kiros2015unifying, karpathy2015deep}.
In particular, \cite{hadsellcvpr2006, salvador2017} consider a loss function that minimizes the distance between pairs of matching cross-modal items while maximizing the distance between non-matching cross-modal pairs.
However, ranking inequality constraints are more naturally expressed by considering triplets composed of a query, a relevant item, and an irrelevant item. This strategy is similar to the Large Margin Nearest Neighbor loss \cite{weinbergerlmnn2009} in which a penalty is computed only if the distance between the query and the relevant item is larger than the distance between the query and the irrelevant item.\\

Our contributions differ from previous work according to three main aspects. First, we propose to model the manifold-alignment issue, which is generally based only on the semantic information \cite{hotelling1936relations,lai2000kernel,bach02jmlr,yan15cvpr}, as a joint learning framework leveraging retrieval and class-based features. In contrast to \cite{salvador2017} which adds an additional classification layer to a manifold-alignment neural model,  we directly integrate semantic information in the loss to refine the  structure of the latent space while also limiting the number of parameters to be learned. 
Second, our model relies on a double-triplet (instead of pairwise learning in \cite{hadsellcvpr2006, salvador2017} or a single triplet as in \cite{weinbergerlmnn2009}) to fit with the joint learning objectives.  Third, we propose a new stochastic gradient descent weighting scheme adapted to such a dual deep embedding architecture, which is computed on minibatches and
automatically performs an adaptive mining of informative triplets.

    \section{AdaMine Deep Learning Model}
       
\begin{figure}[t]
    \begin{center}
    	\includegraphics[width=\linewidth]{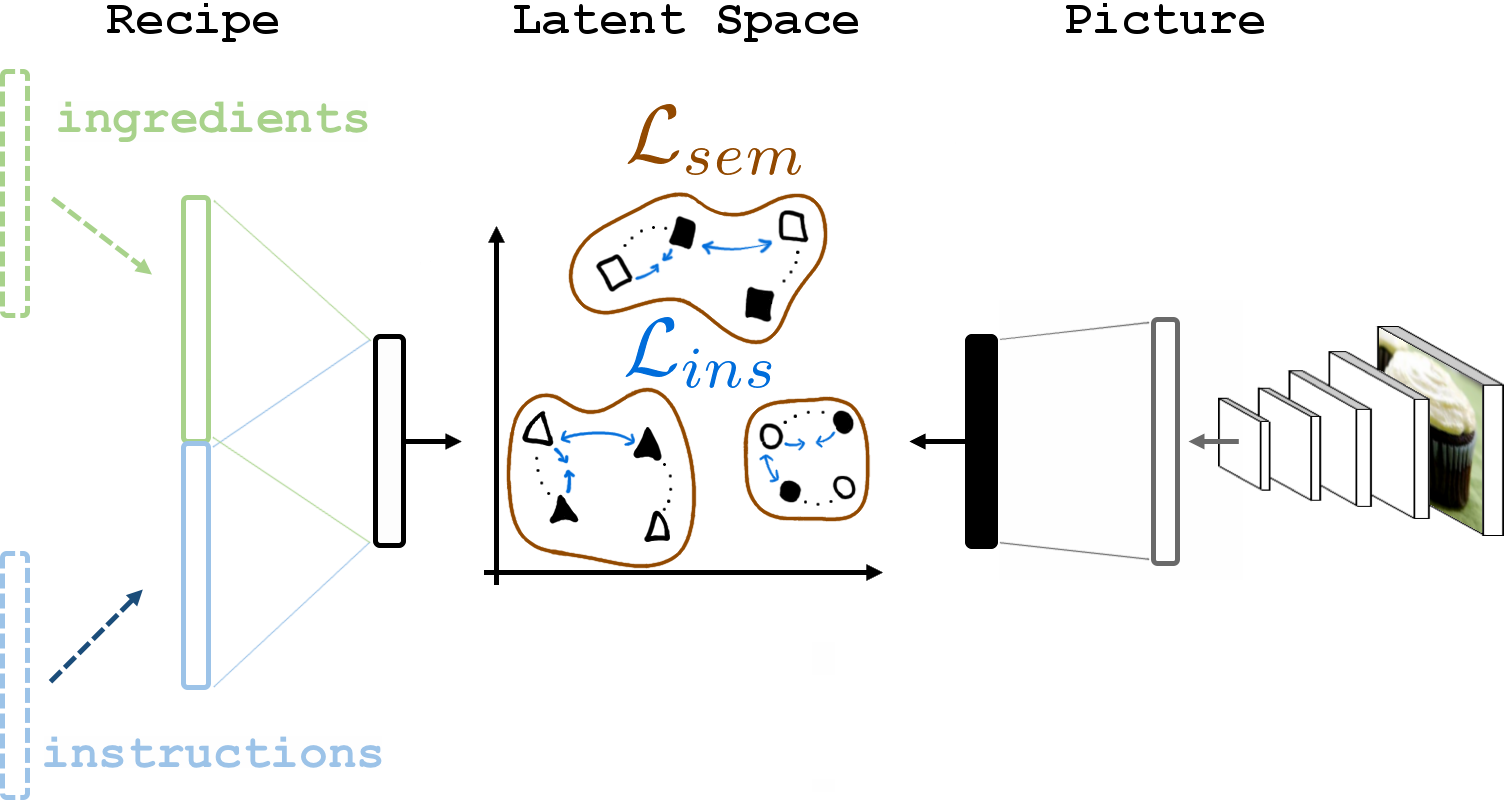}
    	\vspace{-0.4cm}
    	\caption{\small \textbf{AdaMine} overview.}
    	\label{fig:our_architecture}
    	\vspace{-0.6cm}
    \end{center}
\end{figure}

\subsection{Model Overview}
\label{sec:overview}
 The objective of our model \textbf{AdaMine} (ADAptive MINing Embeding) is to learn the representation of recipe items (texts and images) through a joint retrieval and classification learning framework based on a double-triplet learning scheme.
More particularly, our model relies on the following hypotheses:\\
\indent $\bullet$ H1: Aligning items according to a retrieval task allows capturing the fine-grained semantics of items, since the obtained embeddings must rank individual items with respect to each other.\\
\indent $\bullet$ H2: Aligning items according to class meta-data allows  capturing the high-level semantic information underlying items since it ensures the identification of item clusters that correspond to class-based meta-data.\\ 
\indent $\bullet$ H3: Learning simultaneously retrieval and class-based features allows  enforcing a multi-scale structure within the latent space, which covers all aspects of item semantics. In addition, we conjecture that adding a classification layer sequentially to manifold-alignment as in \cite{salvador2017} might be under-effective.\\

Based on these hypotheses, we propose to learn the latent space structure (and item embeddings) by integrating both retrieval objective and semantic information in a single cross-modal metric learning problem (see the Latent Space in Figure \ref{fig:our_architecture}).
We take inspiration from the learning-to-rank retrieval framework by building a learning schema based on query/relevant item/irrelevant item triplets noted  $(x_\iqry, x_\ipos, x_\ineg)$. Following hypothesis H3, we propose a double-triplet learning scheme that relies on both instance-based and semantic-based triplets, noted respectively $(x_\iqry, x_\ipos, x_\ineg)$ and $(x'_\iqry, x'_\ipos, x'_\ineg)$, in order to satisfy the multi-level structure (fine-grained and high-level) underlying semantics. 
More particularly, we learn item embeddings by minimizing the following objective function: 
\begin{align}
    \Loss_{total}(\theta) =  \Loss_{ins}(\theta) + \lambda \Loss_{sem}(\theta)
    \label{eq:tot_loss}
\end{align}
where $\theta$ is the network parameter set. $\Loss_{ins}$ is the loss associated with the retrieval task over instance-based triplets $(x_\iqry, x_\ipos, x_\ineg)$, and $\Loss_{sem}$ is the loss coming with the semantic information over semantic-based triplets $(x'_\iqry, x'_\ipos, x'_\ineg)$. Unlike \cite{salvador2017} that expresses this second term $\Loss_{sem}$ acting as a regularization over the $\Loss_{ins}$ optimization, in our framework, it is expressed as a joint  classification task.

This double-triplet learning framework is a difficult learning problem since the trade-off between $\Loss_{ins}$ and $\Loss_{sem}$ is not only influenced by $\lambda$ but also by the sampling of instance-based and semantic-based triplets and depends on their natural distribution.
Furthermore, the sampling of violating triplets can be difficult as the training progresses which usually leads to vanishing gradient problems that are common in triplet-based losses, and are amplified by our double-triplet framework.
To alleviate these problems, we propose an adaptive sampling strategy that normalizes each loss allowing to fully control the trade-off with $\lambda$ alone while also ensuring non-vanishing gradients throughout the learning process.

In the following, we present the network architecture, each component of our learning framework, and then discuss the learning scheme of our model.

\subsection{Multi-modal Learning Framework}

\subsubsection{Network Architecture}

Our network architecture is based on the proposal of \cite{salvador2017}, which consists of two branches based on deep neural networks that map each modality (image or text recipe) into a common representation space, where they can be compared. Our global architecture is depicted in \autoref{fig:our_architecture}. 

The image branch (top-right part of \autoref{fig:our_architecture}) is composed of a ResNet-50 model \cite{He2015}.  
It contains 50 convolutional layers, totaling more than 25 million parameters.
This architecture is further detailed in~\cite{He2015}, and was chosen in order to obtain comparable results to \cite{salvador2017} by sharing a similar setup.
The ResNet-50 is pretrained on the large-scale dataset of the ImageNet Large Scale Visual Recognition Challenge~\cite{ILSVRC15}, containing 1.2 million images, and is fine-tuned with the whole architecture. 
This neural network is followed by a fully connected layer, which maps the outputs of the ResNet-50 into the latent space, and is trained from scratch.

In the recipe branch (top-left part of \autoref{fig:our_architecture}), ingredients and instructions are first embedded separately, and their obtained representations are then concatenated as input of a fully connected layer that maps the recipe features into the latent space. For ingredients, we use a bidirectional LSTM~\cite{Hochreiter1997} on pretrained embeddings obtained with the word2vec algorithm~\cite{mikolov2013distributed}. With the objective to consider the different granularity levels of  the instruction text, we use a hierarchical LSTM in which the word-level is pretrained using the skip-thought technique~\cite{kiros2015} and is not fine-tuned while the sentence-level is learned from scratch.

\subsubsection{Retrieval loss}
\label{par:baseline_pairwise}
    The objective of the retrieval loss $\Loss_{ins}$ is to learn item embeddings by constraining the latent space according to the following assumptions (Hypothesis H1): 1) ranking items according to a similarity metric so as to gather matching items together and 2) discriminating irrelevant ones.
    We propose to use a loss function $\loss_{ins}$ based on a particular triplet $(x_\iqry, x_\ipos, x_\ineg)$ consisting of a query $x_\iqry$, its matching counterpart in the other modality $x_\ipos$ and a dissimilar item $x_\ineg$. The retrieval loss function $\Loss_{ins}$ is the aggregation of the individual loss $\loss_{ins}$  over all triplets.
    The aim of $\loss_{ins}$ is to provide a fine-grained structure to the latent space where the nearest item from the other modality with respect to the query is optimized to be its matching pair.
    More formally, the individual retrieval loss $\loss_{ins}(\theta, x_\iqry, x_\ipos, x_\ineg)$ is formalized as follows:
    \begin{align}
        \label{eq:loss_tri}
        \loss_{ins}(\theta, x_\iqry, x_\ipos, x_\ineg) &= \left[ d(x_\iqry, x_\ipos) + \alpha - d(x_\iqry, x_\ineg) \right]_{+}
    \end{align}
    where $d(x,y)$ expresses the cosine distance between vectors $x$ and $y$ in the latent space $\FF$.

\subsubsection{Semantic loss}
\label{sec:semantic_triplet}
     $\Loss_{sem}$ is acting as a regularizer capable of taking advantage of semantic information in the multi-modal alignment, without adding extra parameters to the architecture nor graph dependencies.
    To leverage class information (Hypotheses H2), we propose to construct triplets that optimize a surrogate of the k-nearest neighbor classification task. Ideally, for a given query $x_\iqry$, and its corresponding class $c(x_\iqry)$, we want its associated closest sample $x_{\star,\iqry}$ in the feature space to respect $c(x_\iqry) = c(x_{\star,\iqry})$.
    This enforces a semantic structure on the latent space by making sure that related dishes are closer to each other than to non-related ones.
    To achieve this, we propose the individual triplet loss $\loss_{sem}$: 
    \begin{align}
        \label{eq:loss_trisem}
        \loss_{sem}(\theta, x'_\iqry, x'_\ipos, x'_\ineg) &= \left[ d(x'_\iqry, x'_\ipos) + \alpha - d(x'_\iqry, x'_\ineg) \right]_{+}
    \end{align}
    where $x'_\ipos$ belongs to the set 
    of items with the same semantic class $c(x'_\iqry)$ as the query, and $x'_\ineg$ belongs to the set of items with different semantic classes than the one of the query.

    Contrary to the classification machinery adopted by \cite{salvador2017}, $\loss_{sem}$ optimizes semantic relations directly in the latent space  without changing the architecture of the neural network, as shown on \autoref{fig:sem_loss}.
    This promotes a smoothing effect on the space by encouraging instances of the same class to stay closer to each other.

\subsection{Adaptive Learning Schema} 
\label{sec:update_step}
    As commonly used in Deep Learning, we use the stochastic gradient descent (SGD) algorithm which approximates the true gradient over mini-batches. The update term is generally computed by aggregation of the gradient using the \emph{average} over all triplets in the mini-batch. 
    However, this \emph{average} strategy tends to produce a vanishing update with triplet losses. This is especially true towards the end of the learning phase, as the few active constraints are averaged with many zeros coming from the many inactive constraints. We believe this problems is amplified as the size of the training set grows.
       To tackle this issue, our proposed \emph{adaptive} strategy considers an update term $\delta_{adm}$ that takes into account informative triplets only (\textit{i.e.}, non-zero loss).
       More formally, given a mini-batch $\Bat$, $\Pos_q^r$ the set of matching items with respect to a query $x_\iqry$ and $\Pos_q^s$ the set of items with the same class as $x_q$, the update term $\delta_{adm}$ is defined by:
        \begin{align}
            \delta_{adm} = \sum\limits_{x_\iqry \in \Bat} \Bigg( &\sum\limits_{x_\ipos \in \Bat \cap \Pos_q^r} \sum\limits_{x_\ineg \in \Bat \setminus \Pos_q^r} \frac{\nabla\loss_{ins}(\theta,x_\iqry,x_\ipos,x_\ineg)}{\beta_\iret^\prime}\\
            + &\sum\limits_{x_\ipos \in \Bat \cap \Pos_q^s} \sum\limits_{x_\ineg \in \Bat \setminus \Pos_q^s} \lambda\frac{\nabla\loss_{sem}(\theta,x_\iqry,x_\ipos,x_\ineg)}{\beta_\isem^\prime}\Bigg) \nonumber
        \end{align}
        with $\beta_{r}^\prime$ and $\beta_{s}^\prime$ being the number of triplets  contributing to the cost:
        \begin{align}
            \begin{split}
                \beta_\iret^\prime &= \sum\limits_{x_\iqry \in \Bat} \sum\limits_{x_\ipos \in \Bat \cap \Pos_q^r} \sum\limits_{x_\ineg \in \Bat \setminus \Pos_q^r} \mathds{1}_{\loss_{ins} \neq 0}\\
                \beta_\isem^\prime &= \sum\limits_{x_\iqry \in \Bat} \sum\limits_{x_\ipos \in \Bat \cap \Pos_q^s} \sum\limits_{x_\ineg \in \Bat \setminus \Pos_q^s} \mathds{1}_{\loss_{sem} \neq 0}
            \end{split}
        \end{align}

        At the very beginning of the optimization, all triplets contribute to the cost and, as constraints stop being violated, they are dropped. 
        At the end of the training phase, most of the triplets will have no contribution, leaving the hardest negatives to be optimized without vanishing gradient issues. Remark that this corresponds to a curriculum learning starting with the average strategy and ending with the hard negative strategy like in \cite{schroff2015facenet}, but without the burden of finding the time-step at which to switch between strategies as this is automatically controlled by the weights $\beta_r$ and $\beta_s$.
        
         Remark also that an added benefit of $\delta_{adm}$ is due to the independent normalization of each loss by its number of active triplets. Thus $\delta_{adm}$ keeps the trade-off between $\loss_{ins}$ and $\loss_{sem}$ unaffected by difference between the number of active triplets in each loss and allows $\lambda$ to be the only effective control parameter.

    \section{Evaluation protocol}
    \label{sec:exp_setup}

The objective of our evaluation is threefold: 1) Analyzing the impact of our semantic loss that directly  integrates semantic information in the latent space; 2) Testing the effectiveness of our model; 3) Exploring the potential of our model and its learned latent space for solving smart cooking tasks.
All of our experiments are conducted using PyTorch\footnote{http://pytorch.org}, with our own implementation\footnote{https://github.com/Cadene/recipe1m.bootstrap.pytorch} of the experimental setup (\emph{i.e.} preprocessing, architecture and evaluation procedures) described by \cite{salvador2017}. We detail the experimental setup in the following.

\subsection{Dataset}
    We use the  Recipe1M dataset~\cite{salvador2017},  the only large-scale dataset including both English cooking recipes (ingredients and instructions), images, and categories. 
    The raw Recipe1M dataset consists of about 1 million image and recipe pairs.
    It is currently the largest one in English, including twice as many recipes as \cite{kusmierczyk2016understanding} and eight times as many images as \cite{chen2016deep}. Furthermore, the availability of semantic information makes it particularly suited to validate our model: around half of the pairs are associated with a class, among 1048 classes parsed from the~recipe~titles.
    \micael{Using the same preprocessed pairs of recipe-image provided by \cite{salvador2017}}, we end up with 238,399 matching pairs of images and recipes for the training set, while the validation and test sets have 51,119 and 51,303 matching pairs, respectively.

\subsection{Evaluation Methodology}
We carry out a cross-modal retrieval task following the process described in \cite{salvador2017}. Specifically, we first sample 10 unique subsets of 1,000 (1k setup) or 5 unique subsets of 10,000 (10k setup) matching text recipe-image pairs in the test set. Then, we consider each item in a modality as a query (for instance, an image), and we rank items in the other modality (resp. text recipes) according to the cosine distance between the query embedding and the candidate embeddings. The objective is to retrieve the associated item in the other modality at the first rank.
The retrieved lists are evaluated using standard metrics in cross-modal retrieval tasks. For each subset (1k and 10k), we estimate the median retrieval rank (MedR), as well as the recall percentage at top~K (R@K), over all queries in a modality. The R@K corresponds to the percentage of queries for which the matching item is ranked among the top K closest results.

\subsection{Baselines}
    To test the effectiveness of our model \textbf{AdaMine}, we evaluate our multi-modal embeddings  with respect to those obtained by state-of-the-art (SOTA) baselines:\\
\indent $\bullet$ \textbf{CCA}, which denotes the Canonical Correlation Analysis method \cite{hotelling1936relations}. This baseline allows  testing the effectiveness of global alignment methods.\\
\indent $\bullet$ \textbf{PWC}, the pairwise loss with the classification layer from~\cite{salvador2017}. We report their state-of-the-art results for the 1k and 10k setups when available.  This baseline exploits the classification task as a regularization of embedding learning.\\
\indent $\bullet$ \textbf{PWC*}, our implementation of the architecture and loss described by~\cite{salvador2017}. The goal of this baseline is to assess the results of its improved version \textbf{PWC++}, described below.\\
\indent $\bullet$ \textbf{PWC++}, the improved version of our implementation \textbf{PWC*}. More particularly,  we add a positive margin to the pairwise loss adopted in~\cite{salvador2017}, as proposed by~\cite{junlinhucvpr2014}:
            \begin{align}
                \begin{split}
                    \loss_{pw++}( \theta, x_\iqry, x) =~ y &\big[ d(x_\iqry, x) - \alpha_{pos}\big]_{+}\\
                    +~ (1 - y) &\big[ \alpha_{neg} - d(x_\iqry, x)\big]_{+}
                \end{split}
            \end{align}
            with $y=1$ (resp. $y=0$) for pos. (resp. neg.) pairs. The positive margin $\alpha_{pos}$ allows matching pairs to have different representations, thus reducing the risk of overfitting.
            In practice, the positive margin is set to 0.3 and the negative margin to 0.9.\\

We  evaluate  the  effectiveness  of  our model \textbf{AdaMine}, which includes both the triplet loss and the adaptive learning, in different setups, and having the following objectives:\\
\indent $\bullet$ Evaluating the impact of the retrieval loss: we run the \textbf{AdaMine\_{ins}} scenario which  refers to our model with  the instance loss $\mathcal{L}_{ins}$ only and the adaptive learning strategy (the semantic loss $\mathcal{L}_{sem}$ is discarded);\\
\indent $\bullet$ Evaluating the impact of the semantic loss: we run the \textbf{AdaMine\_{sem}} scenario which refers to our model with  the semantic loss $\mathcal{L}_{sem}$ only and the adaptive learning strategy (the instance loss $\mathcal{L}_{ins}$ is discarded);\\
\indent $\bullet$ Evaluating the impact of the strategy used to tackle semantic information: we run the \textbf{AdaMine\_{ins+cls}} scenario which refers to our \textbf{AdaMine} model by replacing the semantic loss by the classification head proposed by \cite{salvador2017};\\
\indent $\bullet$ Measuring the impact of our adaptive learning strategy: we run the \textbf{AdaMine\_{avg}}. The architecture and the losses are identical to our proposal, but instead of using the adaptive learning strategy, this one performs the stochastic gradient descent averaging the gradient over all triplets, as is common practice in the literature;\\
\indent $\bullet$ Evaluating the impact of the text structure: we run our whole model (retrieval and semantic losses + adaptive SGD) by considering either ingredients only (noted \textbf{AdaMine\_{ingr}}) or instructions only (noted \textbf{AdaMine\_{instr}}).

\begin{table*}[ht]
    \centering
    \noindent\makebox[\textwidth]{
        \resizebox{\textwidth}{!}{
            \begin{tabular}{@{\extracolsep{10pt}}lccccccccc@{}}
                \hline
               Scenarios & Strategies& \multicolumn{4}{c}{Image to Textual recipe} & \multicolumn{4}{c}{Textual recipe to Image} \\
                \cline{3-6} \cline{7-10}
                &  & MedR & R@1 & R@5  & R@10 & MedR & R@1 & R@5 & R@10 \\ \hline
                AdaMine\_{ins} & Retrieval loss & 15.4 & 13.3 & 32.1 & 42.6 & 15.8 & 12.3 & 31.1 & 41.7 \\
                AdaMine\_{ins+cls} & Retrieval loss + Classification loss & 14.8 & 13.6 & 32.7 & 43.2 & 15.2 & 12.9 & 31.8 & 42.5 \\
                AdaMine & Retrieval loss + Semantic loss & $\mathbf{13.2}$ & $\mathbf{14.9}$ & $\mathbf{35.3}$ & $\mathbf{45.2}$ & $\mathbf{12.2}$ & $\mathbf{14.8}$ & $\mathbf{34.6}$ & $\mathbf{46.1}$ \\
                \hline
            \end{tabular}
        }
    }
    \caption{\textbf{Impact of the semantic information.} MedR means Median Rank (lower is better). R@K means Recall at K (between 0\% and 100\%, higher is better). The average value over 5 bags of 10,000 pairs each is reported.}
    \label{tab:exp_semantic}
    \vspace{-0.4cm}
\end{table*}

\subsection{Implementation details}

    \paragraph{Network learning.} As adopted by \cite{salvador2017}, we use the Adam~\cite{kingma2014adam} optimizer with a learning rate of $10^{-4}$. Besides, we propose a simpler training scheme: At the beginning of the training phase, we freeze the ResNet-50 weights, optimizing only the text-processing branch, as well as the weights of the mapping of the visual processing branch. After 20 epochs, the weights of the ResNet-50 are unfrozen and the whole architecture is fine-tuned for 60 more epochs.
    For the final model selection, we evaluate the MedR on the validation set at the end of each training epoch, and we keep the model with the best MedR on validation.

    It is worth mentioning that in order to learn our model, a single NVidia Titan X Pascal is used, and the training phase lasts for 30 hours. We also improved the efficiency of the \textbf{PWC} baseline, initially implemented in Torch and requiring 3 days of learning using four NVidia Titan X Pascal to 30 hours on a single NVidia Titan X Pascal. We will release codes for both our model and the \textbf{PWC*} model.
    
\begin{figure}[t]
    \begin{tikzpicture}[x=.8\linewidth,y=0.19cm]
        \def\xmin{0}
        \def\xmax{1.001}
        \def\ymin{12}
        \def\ymax{24}
        \draw[style=help lines, ystep=2, xstep=0.1] (\xmin,\ymin) grid
        (\xmax,\ymax);
        \draw[->] (\xmin,\ymin) -- (\xmax,\ymin) node[right] {$\lambda$};
        \draw[->] (\xmin,\ymin) -- (\xmin,\ymax) node[above] {MedR};
        \node at (0.1, \ymin) [below] {0.1};
        \node at (0.3, \ymin) [below] {0.3};
        \node at (0.5, \ymin) [below] {0.5};
        \node at (0.7, \ymin) [below] {0.7};
        \node at (0.9, \ymin) [below] {0.9};
        \foreach \y in {12,14,...,23}
            \node at (\xmin,\y) [left] {\y};
        \draw[color=blue] plot[smooth,mark=*,mark size=2pt] file {sections/lambda.dat};
    \end{tikzpicture}
    \vspace{-0.2cm}
    \caption{\small MedR scores for different values of the $\lambda$ hyper-parameter, responsible for weighting the semantic regularization cost $\Loss_{sem}$ of \textbf{AdaMine}, calculated over 5 bags of 10.000 validation samples.}
    \label{fig:lambda}
    \vspace{-0.4cm}
\end{figure}
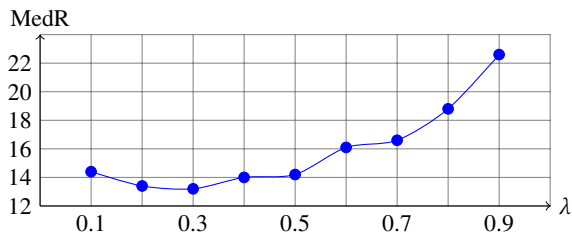

    \paragraph{Parameter choices.} Our model \textbf{AdaMine} is a combination of the adaptive bidirectional instance and semantic triplet losses. Its margin $\alpha$ and the weight $\lambda$ for the semantic cost $\mathcal{L}_{sem}$ are determined using a cross-validation with values varying between 0.1 and 1, and step of 0.1. We finally retained  0.3 for both $\alpha$ and $\lambda$. The parameter $\lambda$ further analyzed in Section 5.1 and in Figure \ref{fig:lambda}.

    \paragraph{Triplet sampling.} As is common with triplet based losses in deep learning, we adopt a per-batch sampling strategy for estimating $\Loss_{ins}$ and $\Loss_{sem}$ (see \autoref{sec:update_step}). The set of multi-modal (image-recipe) matching pairs in the train (resp. validation) set are split in 2383 (resp. 513)  mini-batches of 100  pairs.  Following the dataset structure in which half of the pairs are not labeled by class meta-data, those 100 pairs are split into: 1) 50 randomly selected pairs among those not associated with class information; 2) 50 labeled pairs for which we respect the distribution over all classes in the training set (resp. validation set).
     
    Within each mini-batch, we then build the set of double-triplets fitting with our joint retrieval and semantic loss functions. Each item in the 100 pairs is iteratively seen as the query. The main issue is to build positive and negative sets with respect to this query.     For the retrieval losses, the  item in the other modality associated to the query is assigned to the positive set while the remaining items in the other modality (namely, 99 items) are assigned to the negative instance set. 
    For the semantic loss, we randomly select, as the positive set, one item in the other modality that does not belong to the matching pair while sharing the query class. For the negative set, we consider the remaining items in the other modality that do not belong to the query class. For fair comparison between queries over the mini-batch, we limit the size of the negative sets over each query to the smallest negative ensemble size inside the batch.

    \section{Experiments}
    \label{sec:exps}
    
\begin{figure}[tb]
    \centering
    \subfloat[AdaMine\_{ins}]{
        \includegraphics[width=0.49\linewidth, cfbox=black 0.1pt 0.1pt]{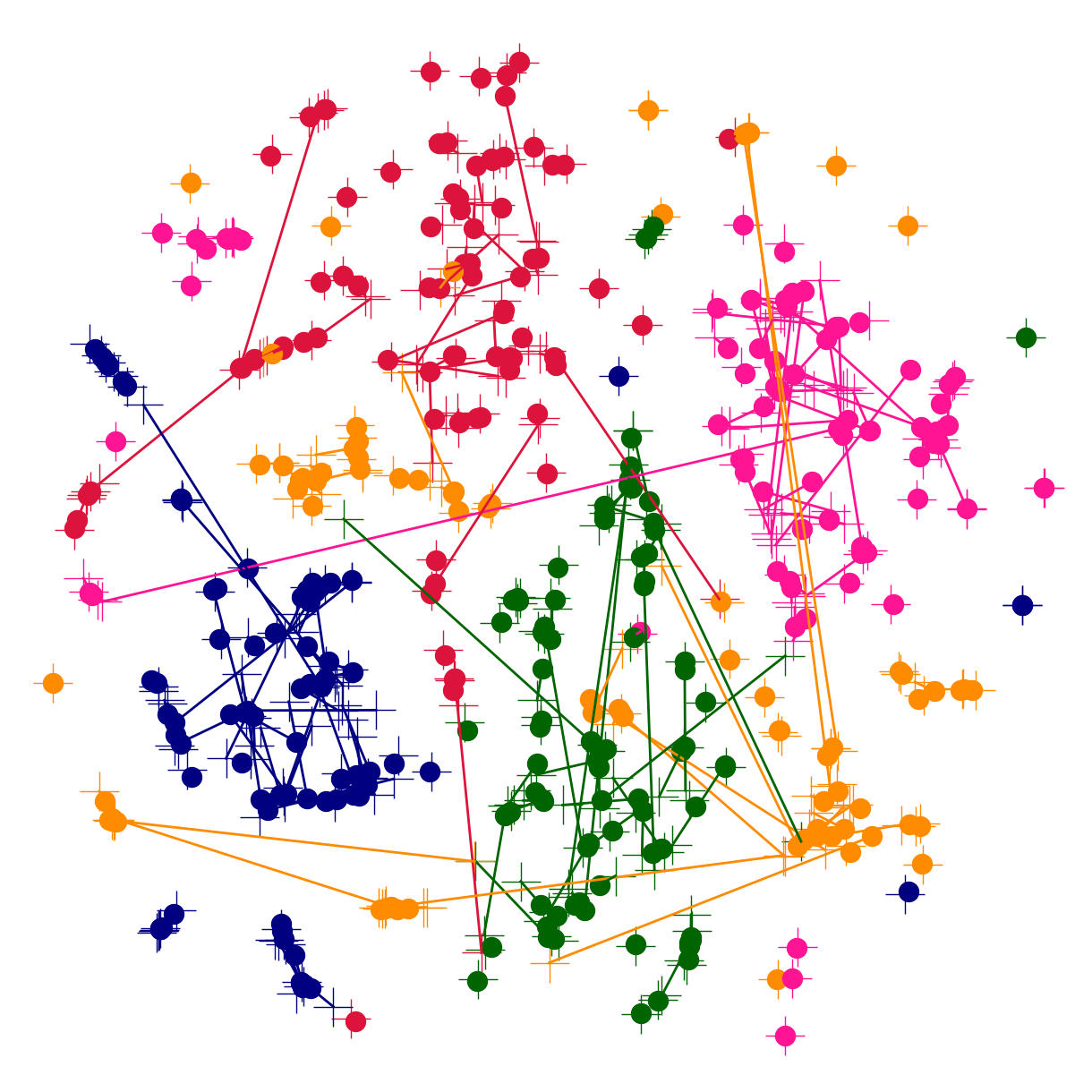}
    }
    \subfloat[AdaMine]{
        \includegraphics[width=0.49\linewidth, cfbox=black 0.1pt 0.1pt]{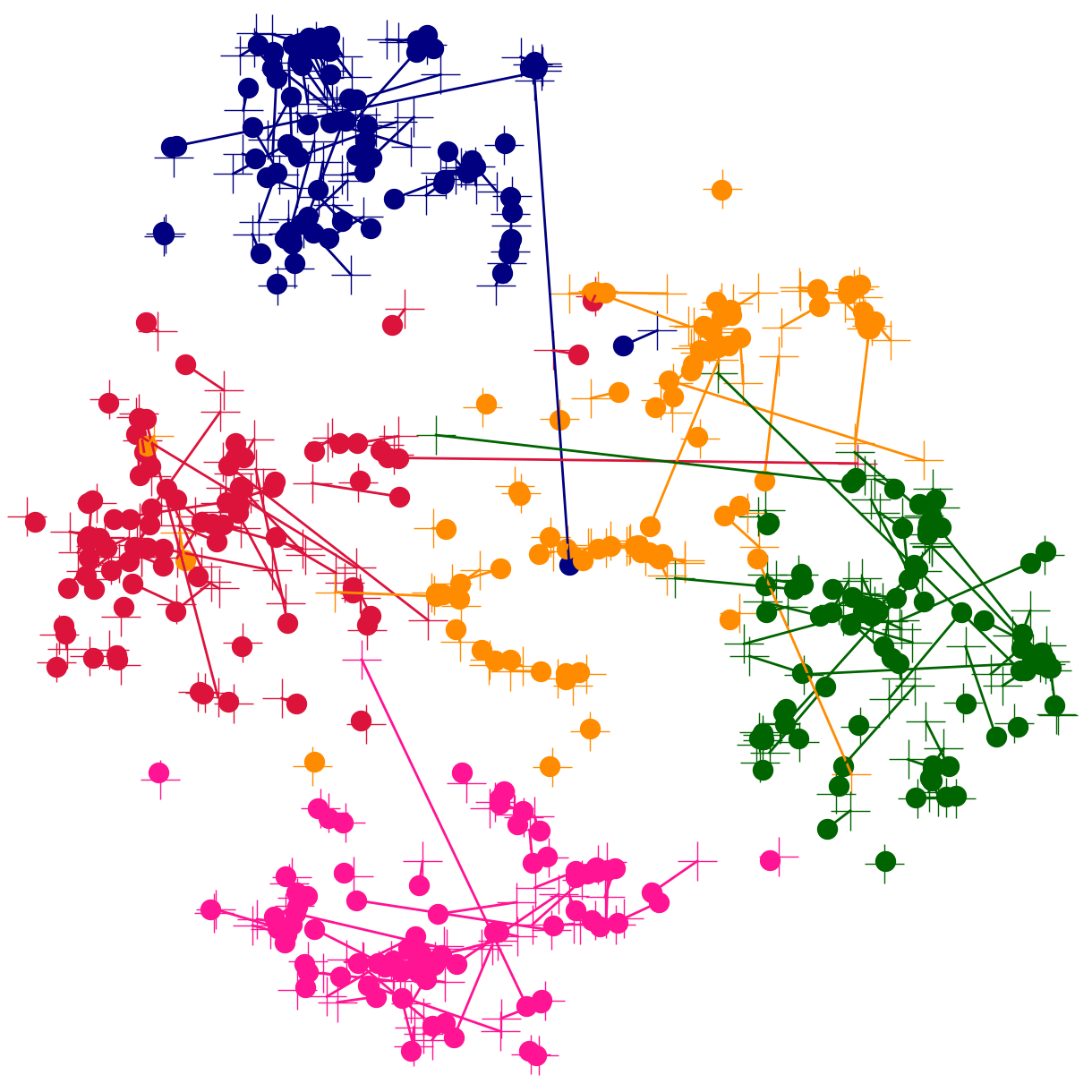}
    }
    \vspace{-0.2cm}
    \caption{\textbf{t-SNE visualization.} Image (resp. Recipe) points are denoted with the $+$ (resp. $\bullet$) symbol. Matching pairs are connected with a trace. Blue points are associated to the cupcake class, orange to hamburger, pink to green beans, green to pork chops, and red to pizza.}
    \label{fig:tsne}
    \vspace{-0.2cm}
\end{figure}

\newcolumntype{C}[1]{>{\centering\let\newline\\\arraybackslash\hspace{0pt}}m{#1}}

\begin{table*}[t]
    \centering
    \begin{tabular}{m{2.9cm}m{5.0cm}m{8cm}@{}m{.4cm}}
        \multicolumn{1}{c}{\textbf{Ingredient query}} & \multicolumn{1}{c}{\textbf{Cooking instruction query}} & \multicolumn{1}{c}{\textbf{Top 5 retrieved images}} \\
        
        \hline
        
        \small \textit{Yogurt, cucumber, salt, garlic clove, fresh mint.} &
        \small \textit{Stir yogurt until smooth.
        Add cucumber, salt, and garlic.
        Garnish with mint.
        Normally eaten with pita bread.
        Enjoy!} &
        
        \raisebox{-.5\height}{\includegraphics[width=\linewidth]{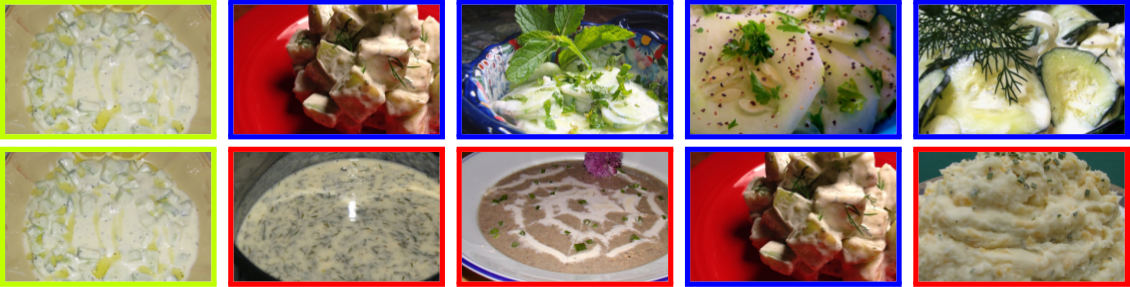}} &
        \begin{sideways}\footnotesize AM\_ins \space\space\space\space\space AM\space\space\space\space\space\space\space\end{sideways}\\
        
        \hline
        
        \small \textit{Olive oil,
        balsamic vinegar,
        thyme,
        lemons, 
        chicken drumsticks with bones and skin,
        garlic, 
        potatoes, 
        parsley.} &
        \small \textit{Whisk together oil, mustard, vinegar, and herbs.
        Season to taste with a bit of salt and pepper and a large pinch or two of brown sugar.
        Place chicken in a non-metal dish and pour marinade on top to coat. [...]} &
        \raisebox{-.5\height}{\includegraphics[width=\linewidth]{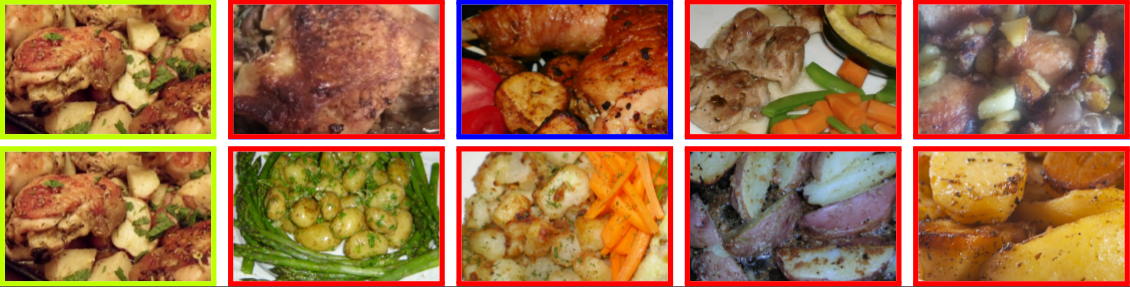}} &
        \begin{sideways}\footnotesize AM\_ins \space\space\space\space\space AM\space\space\space\space\space\space\space\end{sideways}\\
        
        \hline
        
        \small \textit{Pizza dough, hummus, arugula, cherry or grape tomatoes, pitted greek olives, feta cheese.} &
        \small \textit{Cut the dough into two 8-ounce sized pieces.
        Roll the ends under to create round balls.
        Then using a well-floured rolling pin, roll the dough out into 12-inch circles. [...]} &
        \raisebox{-.5\height}{\includegraphics[width=\linewidth]{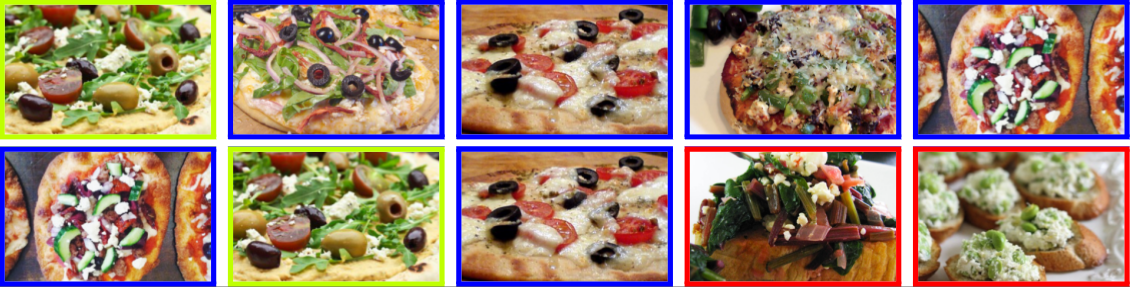}} &
        \begin{sideways}\footnotesize AM\_ins \space\space\space\space\space AM\space\space\space\space\space\space\space\end{sideways}\\
        
        \hline
        
        \small \textit{Unsalted butter, eggs, condensed milk, sugar, vanilla extract, chopped pecans, chocolate chips, butterscotch chips, [...]} &
        \small \textit{Preheat the oven to 375 degrees F.
        In a large bowl, whisk together the melted butter and eggs until combined.
        Whisk in the sweetened condensed milk, sugar, vanilla, pecans, chocolate chips, butterscotch chips, [...]} &
        \raisebox{-.5\height}{\includegraphics[width=\linewidth]{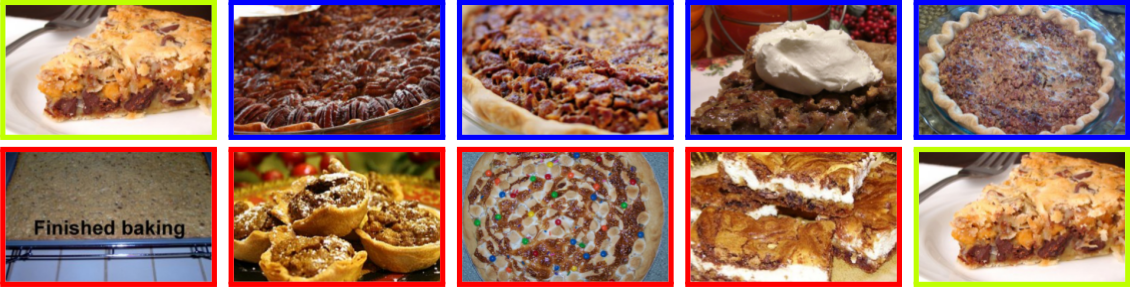}} &
        \begin{sideways}\footnotesize AM\_ins \space\space\space\space\space AM\space\space\space\space\space\space\space\end{sideways}\\
        
    \end{tabular}
    \vspace{\baselineskip}
    \caption{\textbf{Recipe-to-images visualization.} For each recipe, we have the top row, indicating the top 5 images retrieved by our AdaMine model for a given recipe query, and the bottom row, indicating the top 5 images by the triplet loss for the same recipe. In green, the matching image. In blue, images belonging to the same class than the recipe. In red, images belonging to a different class. \textit{AM} indicates AdaMine, and \textit{AM\_ins} AdaMine\_ins.}
    \label{tab:visu2}
    \vspace{-0.4cm}
\end{table*}

\begin{table*}[t]
    \centering
        \begin{tabular}{p{0.1cm}p{0.1cm}ccccc|cccc}
            \hline
            &&& \multicolumn{4}{c|}{Image to Textual recipe} & \multicolumn{4}{c}{Textual recipe to Image} \\ 
            \cline{4-11}
           &&& MedR & R@1 & R@5  & R@10 & MedR & R@1 & R@5 & R@10 \\
            \hline
            \multirow{13}{*}{\begin{sideways}1k items\end{sideways}} &\multirow{6}{*}{\begin{sideways}SOTA\end{sideways}} &Random & 499 & 0.0 & 0.0 & 0.0 & 499 & 0.0 & 0.0 & 0.0 \\
            
            &&CCA \cite{salvador2017} & 15.7 & 14.0 & 32.0 & 43.0 & 24.8 & 9.0 & 24.0 & 35.0 \\
            
            &&PWC \cite{salvador2017} & 5.2 & 24.0 & 51.0 & 65.0 & 5.1 & 25.0 & 52.0 & 65.0 \\
            
            &&PWC \cite{salvador2017}* & $5.0 \pm 0.4$ & $22.8 \pm 1.4$ & $47.7 \pm 1.4$ & $60.1 \pm 1.4$ & $5.3 \pm 0.4$ & $21.2 \pm 1.2$ & $48.0 \pm 1.1$ & $60.4 \pm 1.4$ \\
            
            &&PWC++ & $3.3 \pm 0.4$ & $25.8 \pm 1.6$ & $54.5 \pm 1.3$ & $67.1 \pm 1.4$ & $3.5 \pm 0.5$ & $24.8 \pm 1.1$ & $55.0 \pm 1.8$ & $67.1 \pm 1.2$ \\

            \cline{2-11}
            
            &\multirow{7}{*}{\begin{sideways}Model scenarios\end{sideways}} &AdaMine\_sem & $21.1 \pm 2.0$ & $8.7 \pm 0.7$ & $25.5 \pm 0.9$ & $36.5 \pm 0.9$ & $21.1 \pm 1.9$ & $8.2 \pm 0.9$  & $25.5 \pm 1.0$ & $36.2 \pm 0.9$ \\
            
            &&AdaMine\_ins & $1.5 \pm 0.5$ & 	$37.5 \pm 1.1$ & 	$67.0 \pm 1.3$ & 	$76.8 \pm 1.5$ & 	$1.6 \pm 0.5$ & 	$36.1 \pm 1.6$ & 	$66.6 \pm 1.3$ & 	$76.8 \pm 1.5$\\
            
            &&AdaMine\_ins+cls & $1.1 \pm 0.3$ & 	$38.3 \pm 1.6$ & 	$67.5 \pm 1.2$ & 	$78.0 \pm 0.9$ & 	$1.2 \pm 0.4$ & 	$37.5 \pm 1.4$ & 	$67.7 \pm 1.2$ & 	$77.3 \pm 1.0$\\
            
            &&AdaMine\_avg & $2.3 \pm 0.5$ & 	$30.6 \pm 1.1$ & 	$60.3 \pm 1.2$ & 	$71.4 \pm 1.3$ & 	$2.2 \pm 0.3$ & 	$30.6 \pm 1.8$ & 	$60.6 \pm 1.1$ & 	$71.9 \pm 1.1$\\
            
            &&AdaMine\_ingr  & 	$4.9 \pm 0.5$ & 	$22.6 \pm 1.4$ & 	$48.5 \pm 1.6$ & 	$59.8 \pm 1.3$ & 	$5.0 \pm 0.6$ & 	$21.5 \pm 1.4$ & 	$47.7 \pm 2.1$ & 	$59.8 \pm 1.8$\\
            
            &&AdaMine\_instr & 	$3.9 \pm 0.5$ & 	$24.4 \pm 1.6$ & 	$52.6 \pm 2.0$ & 	$65.4 \pm 1.6$ & 	$3.7 \pm 0.5$ & 	$23.6 \pm 1.7$ & 	$52.7 \pm 1.6$ & 	$65.5 \pm 1.5$\\
            
            \cline{3-11}
            && AdaMine & $\mathbf{1.0 \pm 0.1}$ & $\mathbf{39.8 \pm 1.8}$ & $\mathbf{69.0 \pm 1.8}$ & $\mathbf{77.4 \pm 1.1}$ & $\mathbf{1.0 \pm 0.1}$ & $\mathbf{40.2 \pm 1.6}$ & $\mathbf{68.1 \pm 1.2}$ & $\mathbf{78.7 \pm 1.3}$ \\
            
            \hline
            \hline

            \multirow{8}{*}{\begin{sideways}10k items\end{sideways}} &&PWC++ (best SOTA) & $34.6 \pm 1.0$ & $7.6 \pm 0.2$ & $19.8 \pm 0.1$ & $30.3 \pm 0.4$ & $35.0 \pm 0.9$ & $6.8 \pm 0.2$ & $21.5 \pm 0.2$ & $28.8 \pm 0.3$ \\
            
            \cline{2-11}

            &\multirow{7}{*}{\begin{sideways}Model scenarios\end{sideways}}&AdaMine\_sem & $207.3 \pm 3.9$ & $1.4 \pm 0.3$ & $5.7 \pm 0.3$ & $9.6 \pm 0.3$ & $205.4 \pm 3.2$ & $1.4 \pm 0.1$  & $5.4 \pm 0.2$ & $9.1 \pm 0.4$ \\
            
            &&AdaMine\_ins & $15.4 \pm 0.5$ & $13.3 \pm 0.2$ & $32.1 \pm 0.7$ & $42.6 \pm 0.8$ & $15.8 \pm 0.7$ & $12.3 \pm 0.3$ & $31.1 \pm 0.5$ & $41.7 \pm 0.6$ \\
            
            &&AdaMine\_ins+cls & $14.8 \pm 0.4$ & $13.6 \pm 0.2$ & $32.7 \pm 0.4$ & $43.2 \pm 0.3$ & $15.2 \pm 0.4$ & $12.9 \pm 0.3$ & $31.8 \pm 0.3$ & $42.5 \pm 0.2$ \\
            
            &&AdaMine\_avg & $24.6 \pm 0.8$ & $10.0 \pm 0.2$ & $25.9 \pm 0.4$ & $35.7 \pm 0.5$ & $24.0 \pm 0.6$ & $9.2 \pm 0.4$ & $25.4 \pm 0.5$ & $35.3 \pm 0.4$\\

            &&AdaMine\_ingr & 	$52.8 \pm 1.2$ & 	$6.5 \pm 0.2$ & 	$17.9 \pm 0.2$ & 	$25.8 \pm 0.3$ & 	$53.8 \pm 0.7$ & 	$5.8 \pm 0.3$ & 	$17.3 \pm 0.2$ & 	$25.0 \pm 0.2$\\
            
            &&AdaMine\_instr & 	$39.0 \pm 0.9$ & 	$6.4 \pm 0.1$ & 	$18.9 \pm 0.4$ & 	$27.6 \pm 0.5$ & 	$39.2 \pm 0.7$ & 	$5.7 \pm 0.4$ & 	$17.9 \pm 0.6$ & 	$26.6 \pm 0.5$\\
            \cline{3-11}
            &&AdaMine & $\mathbf{13.2 \pm 0.4}$ & $\mathbf{14.9 \pm 0.3}$ & $\mathbf{35.3 \pm 0.2}$ & $\mathbf{45.2 \pm 0.2}$ & $\mathbf{12.2 \pm 0.4}$ & $\mathbf{14.8 \pm 0.3}$ & $\mathbf{34.6 \pm 0.3}$ & $\mathbf{46.1 \pm 0.3}$ \\
            \hline
        \end{tabular}
    \vspace{\baselineskip}
    \caption{\textbf{State-of-the-art comparison.} MedR means Median Rank (lower is better). R@K means Recall at K (between 0\% and 100\%, higher is better). The mean and std values over 10 (resp. 5) bags of 1k (resp. 10k) pairs each are reported for the top (resp. bottom) table. Items marked with a star (*) are our reimplementation of the cited methods.}
    \label{tab:exp1}
    \vspace{-0.4cm}
\end{table*}
\subsection{Analysis of the semantic contribution}
    We analyze  our main hypotheses related to the importance of  semantic information for learning multi-modal embeddings (see Hypothesis H2 in \ref{sec:overview}). Specifically, in this part we test whether semantic information can help to better structure the latent space, taking into account class information and imposing structural coherence. Compared with \cite{salvador2017} which adds an additional classification layer, we believe that directly injecting this semantic information with a global loss $\mathcal{L}(\theta)$ (\autoref{eq:tot_loss}) comes as a more natural approach to integrating class-based meta-data (see Hypothesis H3 in \ref{sec:overview}).
    
    To test this intuition, we start by quantifying, in \autoref{tab:exp_semantic}, the impacts of the semantic information in the learning process. To do so, we evaluate the effectiveness of different scenarios of our model \textbf{AdaMine} with respect to the multi-modal retrieval task (image-to-text and text-to-image) in terms of \emph{MedR} and Recall at ranks 1, 5, and 10.
    Compared with a retrieval loss alone (\textbf{AdaMine\_{ins}}), we point out that adding semantic information with a classification cost \textbf{AdaMine\_{ins+cls}} or a semantic loss \textbf{AdaMine}  improves the results. 
    When evaluating with 10,000 pairs (10k setting), while \textbf{AdaMine\_{ins}} obtains MedRs 15.4 and 15.8, the semantic models (\textbf{AdaMine\_{ins+cls}} and \textbf{AdaMine}) lower these values to 14.8 and 15.2, and 13.2 and 12.2, respectively (lower is better) for both retrieval tasks (image-to-text and text-to-image).
    
    The importance of semantic information becomes clearer when we directly compare the impact of adding the semantic loss to the base model (\textbf{AdaMine} vs \textbf{AdaMine\_{ins}}), since the former obtains the best results for every metric.
    To better understand this phenomenon, we depict in \autoref{fig:tsne} item embeddings obtained by the \textbf{AdaMine\_{ins}} and \textbf{AdaMine} models using a t-SNE visualization.
    This figure is generated by selecting 400 matching recipe-image pairs (800 data points), which are randomly selected from, and equally distributed among 5 of the most occurring classes of the Recipe1M dataset.
    Each item is colored according to its category (e.g., blue points for the cupcake class), and items of the same instance are connected with a trace. Therefore, \autoref{fig:tsne} allows drawing two conclusions:
    1) our model---on the right side of the figure---is able to structure the latent space while keeping items of the same class close to each other (see color clusters);
    2) our model reduces the sum of distances between pairs of instances (in the figure, connected with traces), thus reducing the MedR and increasing the recall.
     We also illustrate this comparison through qualitative examples. In \autoref{tab:visu2}, \textbf{AdaMine} (top row) and \textbf{AdaMine\_{ins}} (bottom row) are compared on four queries, for which both models are able to rank the correct match in the top-5 among 10,000 candidates.
    For the first and second queries (cucumber salad and roasted chicken, respectively), both models are able to retrieve the matching image in the first position. However, the rest of the top images retrieved by our model are semantically related to the query, by sharing critical ingredients (cucumber, chicken) of the recipe. 
    In the third and fourth queries (pizza and chocolate chip, respectively), our model is able to rank both the matching image and semantically connected samples in a more coherent way, due to a better alignment of the retrieval space produced by the semantic modeling.
    These results reinforce our intuition that it is necessary to integrate semantic information in addition to item pairwise anchors while learning multi-modal embeddings.\\
    
    Second, we evaluate our intuition that classification is under-effective for integrating the semantics within the latent space (see Hypothesis H3 in \ref{sec:overview}). \autoref{tab:exp_semantic} shows that our semantic loss \textbf{AdaMine}, proposed in \autoref{sec:semantic_triplet}, outperforms our model scenario \textbf{AdaMine\_{ins+cls}} which relies on a classification head as proposed~in~\cite{salvador2017}. For instance, we obtain an improvement of $+9.57\%$ in terms of $R@1$ with respect to the classification loss setting \textbf{AdaMine\_{ins+cls}}. This result suggests that our semantic loss is more appropriate to organize the latent space so as to retrieve text-image matching pairs. 
    It becomes important, then, to understand the impacts of the weighting factor $\lambda$ between the two losses $\Loss_{ins}$ and $\Loss_{sem}$ (\autoref{eq:tot_loss}). In \autoref{fig:lambda}, we observe a fair level of robustness for lower values of $\lambda$, but any value over 0.5 has a hindering effect on the retrieval task, since the semantic grouping starts to be of considerable importance. 
    These experiments confirm the importance of additional semantic clues: despite having one million less parameters than \cite{salvador2017}'s proposal, our approach still achieves better scores, when compared to the addition of the classification head.

 \begin{table*}[t]
        \centering
        \begin{tabular}{c c c c c}
            
          Mushrooms & Pineapple & Olives & Pepperoni & Strawberries \\
        
          \includegraphics[width=0.17\linewidth]{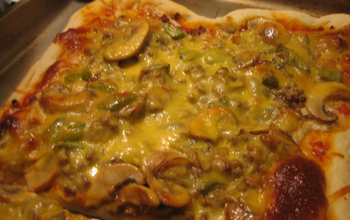} &
          \includegraphics[width=0.17\linewidth]{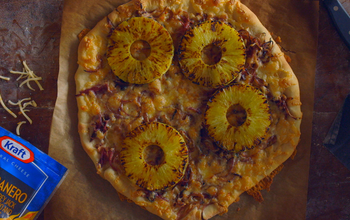} &
          \includegraphics[width=0.17\linewidth]{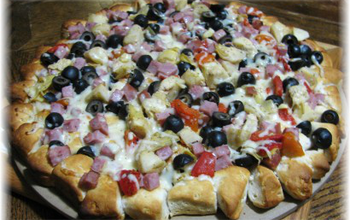} &
          \includegraphics[width=0.17\linewidth]{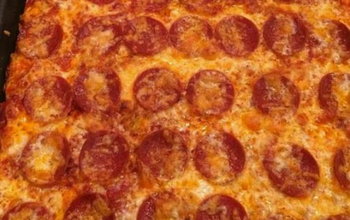} &
          \includegraphics[width=0.17\linewidth]{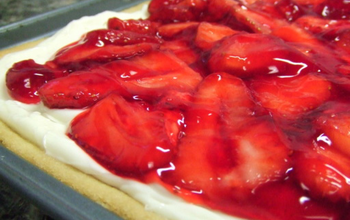} \\
          
          \includegraphics[width=0.17\linewidth]{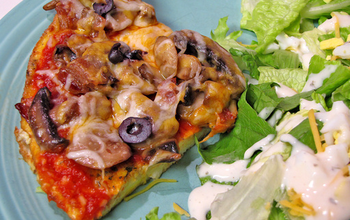} &
          \includegraphics[width=0.17\linewidth]{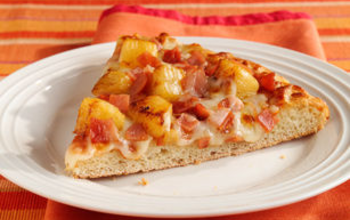} &
          \includegraphics[width=0.17\linewidth]{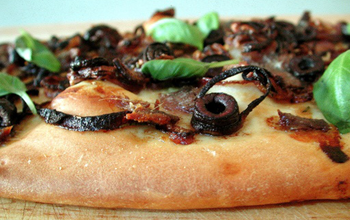} &
          \includegraphics[width=0.17\linewidth]{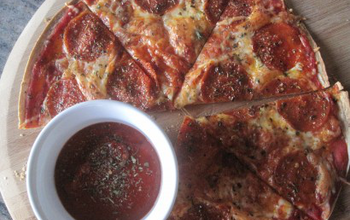} &
          \includegraphics[width=0.17\linewidth]{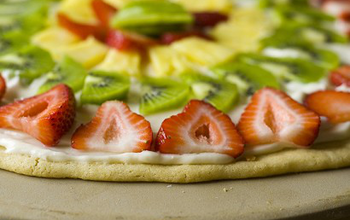}
        \end{tabular}
        \vspace{\baselineskip}
        \vspace{-0.2cm}
        \caption{Ingredient-to-Image task. Examples of images in the top 20 results when searching for an ingredient within the class \emph{Pizza}.}
        \label{tab:visu_ingr_to_image_pizza}
        \vspace{-0.4cm}
    \end{table*}  
    
    \begin{table*}[t]
        \centering
        \begin{tabular}{m{0.28\linewidth}|c c}
            \centering
            Ingredients and Instructions Query
            &
            \multicolumn{1}{c}{
                Top 4 retrieved images
            }\\
            
            \hline
            \begin{minipage}{\linewidth}
                \small \begin{center}\textit{Oregano, Zucchini, Tofu, Bell pepper,}\end{center}
                 \begin{center}\textit{Onions, \st{\textbf{Broccoli}}, Olive Oil}\end{center}
                 \vspace{0.5em}
                 1. Cut all ingredients into small pieces.\\
                \st{2. Put \textbf{broccoli} in hot water for 10 min}\\
                3. Heat olive oil in pan and put oregano in it.\\
                4. Put cottage cheese and saute for 1 minute.\\
                \st{5. Put onion, bell pepper, \textbf{broccoli}, zucchini.}\\
                6. Put burnt chilli garlic dressing with salt.\\
                7. Saute for 1 minutes.
            \end{minipage}
            &
            \multicolumn{1}{c}{
                    \begin{minipage}{.64\linewidth}
                        \vspace{0.1em}
                        \includegraphics[width=0.246\linewidth]{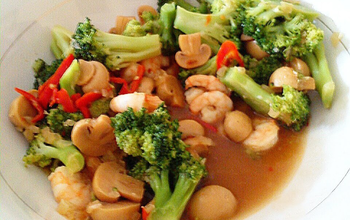}
                        \includegraphics[width=0.246\linewidth]{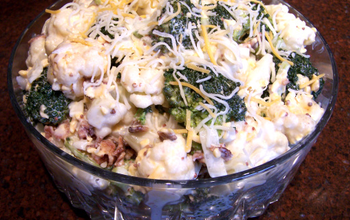}
                        \includegraphics[width=0.246\linewidth]{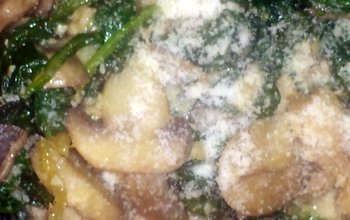}
                        \includegraphics[width=0.246\linewidth]{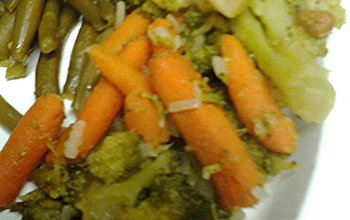} \\
                        
                        \vspace{-0.7em}
                        \includegraphics[width=0.246\linewidth]{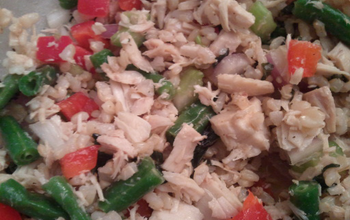}
                        \includegraphics[width=0.246\linewidth]{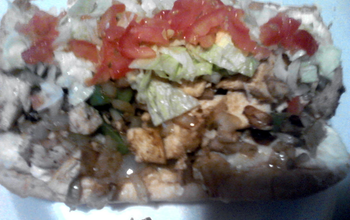}
                        \includegraphics[width=0.246\linewidth]{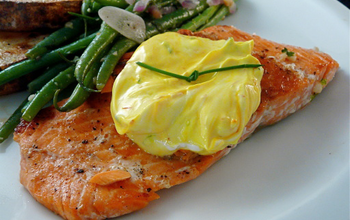}
                        \includegraphics[width=0.246\linewidth]{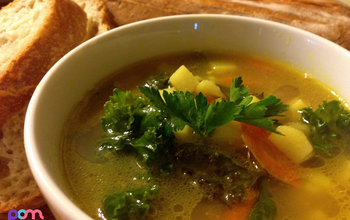}
                    \end{minipage}
            }
            &
            \multicolumn{1}{c}{
                    \begin{minipage}{.02\linewidth}
        \begin{sideways}\footnotesize  with broccoli \space\space\space\space\space\space\end{sideways}\\
        
        \begin{sideways}\footnotesize \space\space\space\space without broccoli \space\space \end{sideways}
                    \end{minipage}
            }
        \end{tabular}
        \vspace{\baselineskip}
        \vspace{-0.2cm}
        \caption{Removing ingredient task. Top 4 retrieved images with (top row) and without (bottom row)  broccoli in the ingredient and instruction lists w.r.t the original recipe (Tofu Saut\'e).}
        \label{tab:vvisu_remove_ingrs_removed}
        \vspace{-0.6cm}
    \end{table*}

\subsection{Testing the effectiveness of the model}
    In the following, we evaluate the effectiveness of our model, compared to different baseline models. Results are presented in \autoref{tab:exp1} for both image-to-text and text-to-image retrieval tasks. We report results on the 1K setup and test the robustness of our model on the 10k setup by reporting only the best state-of-the-art (SOTA) baseline for comparison.
    From a general point of view, we observe that our model \textbf{AdaMine} overpasses the different baselines and model scenarios. Small values of standard deviation outlines the low variability of experimented models, and accordingly the robustness of obtained results. For instance, our model reaches a value equal to 1 for the Median Rank metric (MedR) for the 1k setting and both retrieval tasks while the well-known  SOTA models CCA and PWC++ obtain respectively 15.7 and 3.3. \micael{Contrary to PWC, all of our model scenarios, denoted \textbf{AdaMine\_$_{*}$}, adopt the triplet loss.} Ablation tests on our proposals show their effectiveness. This trend is noticed over all retrieval tasks and all metrics. The comparison of the results obtained over  1k and 10k settings outlines the same statement with larger improvements (with similar standard deviation) for our model \textbf{AdaMine} with respect to SOTA models and \textbf{AdaMine}-based scenarios. 
    More particularly, we first begin our discussion with the comparison with respect to SOTA models and outline  the following statements:
    
    $\bullet$ Global alignment models (baseline \textbf{CCA}) are less effective than advanced models (\textbf{PWC}, \textbf{PWC++}, and \textbf{AdaMine}). Indeed, the \textbf{CCA} model obtains a MedR value of 15.7 for the image-to-text retrieval task (1k setting) while the metric range of advanced models is between 1 and 5.2. This suggests the effectiveness of taking into account dissimilar pairs during the learning process.

    $\bullet$ We observe that our triplet based model \textbf{AdaMine} consistently outperforms pairwise methods (\textbf{PWC} and \textbf{PWC++}). For instance, our model obtains a significant decrease of $-61.84\%$ in terms of MedR with respect to \textbf{PWC++} for the 10k setting and the image-to-text retrieval task. This suggests that relative cosine distances are better at structuring the latent space than absolute cosine distances.
    
    $\bullet$ Our model \textbf{AdaMine} surpasses the current state-of-the-art results by a large margin. For the 1k setup, it reduces the medR score by a factor of 5---from 5.2 and 5.1 to 1.0 and 1.0---, and by a factor bigger than 3 for the 10k setup. One strength of our model is that it has fewer parameters than \textbf{PWC++} and \textbf{PWC}, since the feature space is directly optimized with a semantic loss, without the addition a parameter-heavy head to the model.     
    
    Second, the comparison according to different versions of our model outlines three main statements: 
    
    $\bullet$ The analysis of \textbf{AdaMine\_ins}, \textbf{AdaMine\_ins+cls}, and \textbf{AdaMine} corroborates  the results observed in Section 5.1 dealing with the impact of the  semantic loss on the performance of the model. In the 1k setting, the instance-based approach (\textbf{AdaMine\_ins}) achieves a MedRs value equal of 1.5 and 1.6 for both tasks (lower is better), while the addition of a classification head (\textbf{AdaMine\_{ins+cls}}), proposed by \cite{salvador2017}, improves these results to 1.1  and 1.2. Removing the classification head and adding a semantic loss (\textbf{AdaMine}) further improves the results to 1 for both retrieval tasks which further validates Hypothesis H3 in \ref{sec:overview}.

    $\bullet$ The adaptive sampling strategy described in \autoref{sec:update_step} strongly contributes to the good results of \textbf{AdaMine}. With \textbf{AdaMine\_avg}, we test the same setup of \textbf{AdaMine}, replacing the adaptive strategy with the average one. The importance of removing triplets that are not contributing to the loss becomes evident when the scores for both strategies are compared: 24.6 and 24.0 of MedR (lower is better) for \textbf{AdaMine\_avg}, and 13.2 and 12.2 for \textbf{AdaMine}, an improvement of roughly 46.34\% and 49.17\%.

    $\bullet$ \textbf{AdaMine} combines the information coming from the image and all the parts of the recipe (instructions and ingredients), attaining high scores. When compared to the degraded models \textbf{AdaMine\_ingr} and \textbf{AdaMine\_instr}, we conclude that both textual information are complementary and necessary for correctly identifying the recipe of a plate. While \textbf{AdaMine} achieves MedRs of 13.2 and 12.2 (lower is better), the scenarios without instructions or without ingredients achieve 52.8 and 53.8, and 39.0 and 39.2, respectively.

\subsection{Qualitative studies on downstream tasks}

    In what follows, we discuss the potential of our model for promising cooking-related application tasks. We particularly focus on downstream tasks in which the current setting might be applied. We provide illustrative examples issued from the testing set of our evaluation process.
    For better readability, we always show the results as images, even for text recipes for which we display their corresponding~original~picture.

    \paragraph{Ingredient To Image}
    An interesting ability of our model is to map ingredients into the latent space. One example of task is  to retrieve recipes containing specific ingredients that could be visually identified.
    This is particularly useful when one would like to know what they can cook using aliments available in their fridge.     To demonstrate this process, we create each recipe query as follows: 1) for the ingredients part, we use a single word which corresponds to the ingredient we want to retrieve; 2) for the instructions part, we use the average of the instruction embeddings over all the training set. Then, we project our query into the multi-modal space and retrieve the nearest neighbors among 10,000 images randomly picked from the testing set. 
    We show on \autoref{tab:visu_ingr_to_image_pizza} examples of retrieved images when searching for different ingredients while constraining the results to the class \emph{pizza}. Searching for \emph{pineapple} or \emph{olives} results in different types of pizzas. An interesting remark is that searching for \emph{strawberries} inside the class \emph{pizza} yields images of \emph{fruit pizza} containing strawberries, \textit{i.e.}, images that are visually similar to pizzas while containing the required ingredient. This shows the fine-grain structure of the latent space in which recipes and images are organized by visual or semantic similarity inside the different classes.

    \paragraph{Removing ingredients}

    The capacity of finely model the presence or absence of specific ingredients may be interesting for generating menus, specially for users with dietary restrictions (for instance, peanut or lactose intolerance, or vegetarians and vegans).
    To do so, we randomly select a recipe having \emph{broccoli} in its ingredients list (\autoref{tab:vvisu_remove_ingrs_removed}, first column) and retrieve the top 4 closest images in the embedding space from 1000 recipe images (\autoref{tab:vvisu_remove_ingrs_removed}, top row). Then we remove the \emph{broccoli} in the ingredients and remove the instructions having the \emph{broccoli} word. Finally, we retrieve once again the top 4 images associated to this "modified" recipe (\autoref{tab:vvisu_remove_ingrs_removed}, bottom row).
    The retrieved images using the original recipe have broccoli, whereas the retrieved images using the modified recipe do not have broccoli. This reinforces our previous statement, highlighting the ability of our latent space to correctly discriminate items with respect to ingredients.

    \section{Conclusion}

In this paper, we introduce the AdaMine approach for learning crossmodal embeddings in the context of a large-scale cooking oriented retrieval task (image to recipe, and \textit{vice versa}). Our main contribution relies on a joint retrieval and classification learning framework in which semantic information is directly injected in the cross-modal metric learning. This allows refining the multi-modal latent space by limiting the number of parameters to be learned. For learning our double-triplet learning scheme, we propose an adaptive strategy for informative triplet mining. 
AdaMine is evaluated on the very large scale and challenging Recipe1M crossmodal dataset, outperforming the state-of-the-art models. We also outline the benefit of incorporating semantic information and show the quality of the learned latent space with respect to downstream tasks. We are convinced that such very large scale multimodal deep embeddings frameworks offer new opportunities to explore joint combinations of Vision and Language understanding.
Indeed, we plan in future work to extend our model by considering hierarchical levels within object semantics to better refine the structure of the latent space.
    
    \section*{Acknowledgment}
    This work was partially supported by CNPq --- Brazilian's National Council for Scientific and Technological Development --- and by Labex SMART, supported by French state funds managed by the ANR within the Investissements d'Avenir program under reference ANR-11-LABX-65.
    
    {
        \bibliographystyle{ACM-Reference-Format}
        \bibliography{egbib}
    }
    
    \end{sloppypar}
\end{document}